\documentclass[runningheads]{llncs}

\makeatletter
\def\subsubsection{%
    \@startsection
    {subsubsection}                   
    {3}                               
    {0ex}                             
    {2.5ex}                             
    {2.5ex}                            
    {\normalfont\normalsize\bfseries}
}
\makeatother

\usepackage{graphicx}
\usepackage{marvosym}
\usepackage{amsmath,amssymb,amsfonts}
\usepackage{algorithmic}
\usepackage{graphicx}
\usepackage{textcomp}
\usepackage[T5]{fontenc}
\usepackage[utf8]{inputenc}
\usepackage{pifont}
\usepackage{float}
\usepackage{caption}
\usepackage{placeins}
\usepackage{array}
\newcolumntype{P}[1]{>{\centering\arraybackslash}p{#1}}
\newcolumntype{M}[1]{>{\centering\arraybackslash}m{#1}}
\usepackage{multirow}
\usepackage[bookmarks,bookmarksopen,bookmarksdepth=3, bookmarksnumbered]{hyperref}
\hypersetup{colorlinks, citecolor=blue, linkcolor=blue, urlcolor=blue}

\begin{document}
\title{Vietnamese Word Segmentation\\with SVM: Ambiguity Reduction\\and Suffix Capture}

\titlerunning{Vietnamese Word Segmentation with SVM}

\author{Duc-Vu Nguyen\inst{1,3}\textsuperscript{(\Letter)}\and
Dang Van Thin\inst{1,3} \and \\Kiet Van Nguyen\inst{2,3} \and Ngan Luu-Thuy Nguyen\inst{2,3}}

\authorrunning{D.-V. Nguyen et al.}

\institute{Multimedia Communications Laboratory, University of Information Technology,\\Ho Chi Minh City, Vietnam \\
\email{\{vund,thindv\}@uit.edu.vn}\\ \and
University of Information Technology, Ho Chi Minh City, Vietnam\\
\email{\{kietnv,ngannlt\}@uit.edu.vn} \and Vietnam National University, Ho Chi Minh City, Vietnam}

\maketitle

\begin{abstract}

In this paper, we approach Vietnamese word segmentation as a binary classification by using the Support Vector Machine classifier. We inherit features from prior works such as n-gram of syllables, n-gram of syllable types, and checking conjunction of adjacent syllables in the dictionary. We propose two novel ways to feature extraction, one to reduce the overlap ambiguity and the other to increase the ability to predict unknown words containing suffixes. Different from UETsegmenter and RDRsegmenter, two state-of-the-art Vietnamese word segmentation methods, we do not employ the longest matching algorithm as an initial processing step or any post-processing technique. According to experimental results on benchmark Vietnamese datasets, our proposed method obtained a better $\text{F}_\text{1}\text{-score}$ than the prior state-of-the-art methods UETsegmenter, and RDRsegmenter.

\end{abstract}
\section{Introduction} \label{introduction}
Word segmentation is an essential task in Vietnamese natural language processing, which has a significant impact on higher processing levels \cite{diend1,ha2003,nguyenetal2006}. Unlike English, white spaces in Vietnamese written text can function as a syllable separator or a word separator. For example, the Vietnamese string ``hiện đại hóa đất nước'' ($\text{modernize}_\text{hiện\_đại\_hoá}$ $\text{country}_\text{đất\_nước}$), which consists of five syllables, is segmented into ``hiện\_đại\_hoá đất\_nước''. Underscores denote the white spaces which function as syllable separator, and white spaces are used for word separation. Vietnamese word segmentation can be considered as a binary classification problem with two classes: underscore and white-space \cite{thainp1}.

Vietnamese is an isolated language and every Vietnamese word has exactly one form \cite{hongphuong08}. Vietnamese words are constituted by one or more syllables. According to the statistics reported in \cite{hongphuong08}, and \cite{phongnt1}, about 16\% of Vietnamese words are single-syllable words and 71\% are two-syllable words. Single-syllable words account for about 81\% of Vietnamese syllables, which means 19\% syllables are not meaningful when standing alone. The string ``loại hình phạt'' (3 syllables) can be segmented as ``loại\_hình phạt'' ($\text{type}_\text{loại\_hình}$ $\text{penalize}_\text{phạt}$) or ``loại hình\_phạt'' ($\text{type}_\text{loại}$ $\text{penalty}_\text{hình\_phạt}$). This phenomenon is called ``overlap ambiguity involving three consecutive syllables'' by the authors in \cite{hongphuong08}. All of the above have created challenges in Vietnamese word segmentation \cite{quynt1}.

We have an observation that solving overlap ambiguity is essential for the Vietnamese word segmentation task. The authors in \cite{hongphuong08} proposed the ambiguity resolver, which uses a bi-gram language model. Their proposal has slightly improved the Vietnamese word segmentation result. Additionally, the binary classifier for the Vietnamese word segmentation trained by the authors in \cite{phongnt1} still causes overlap ambiguity cases. They used rules based on the dictionary and threshold for the classifier in the post-processing phase to handle overlap ambiguities. Experimental results on the benchmark Vietnamese treebank show that the approach of the authors in \cite{phongnt1} outperforms the previous state-of-the-art method of the authors in \cite{hongphuong08}. Therefore, we decided to inspire the idea from the authors in \cite{phongnt1} in handling overlap ambiguities. However, we have assumed how the performance of our method changes when using feature templates to reduce overlap ambiguity cases without post-processing.

From a different point of view, the authors in \cite{nghiem2008} proposed affixes features as a part of the rich feature set in their Vietnamese POS tagging method. Additionally, the authors in \cite{lehongtag10} utilized potential affixes to improve the performance of unknown words (accuracy of 80.69\% on Vietnamese POS tagging task of Vietnamese treebank \cite{thainp1}). In practice, we can not perform part-of-speech (POS) tagging for unknown words if these unknown words can not be constituted by machine annotated word segmentation. Therefore, we decide to study the impact of affixes on the performance of word segmentation. We approach Vietnamese word segmentation with a uni-directional model in which labels are predicted from left to right of a sentence based on a syllable window. Because those labels from the left hand have been predicted, we can utilize information of suffixes to improve Vietnamese word segmentation.

In this paper, we propose a feature-based method using SVM classifier to solve the Vietnamese word segmentation task. Our method considers Vietnamese word segmentation as a binary classification with two classes: underscore and white-space \cite{phongnt1}, in which a majority of feature templates are inherited from the research of the authors in \cite{nguyenetal2006,phongnt1}. Two novel feature templates in our method are to reduce ambiguity cases and capture unknown words containing suffixes. Our proposed method obtained better $\text{F}_\text{1}\text{-score}$ than the previous state-of-the-art methods JVnSegmenter \cite{nguyenetal2006}, vnTokenizer \cite{hongphuong08}, DongDu \cite{luu2012}, UETsegmenter \cite{phongnt1}, and RDRsegmenter \cite{datnq1} measured on the Vietnamese treebank \cite{thainp1} for Vietnamese word segmentation task. Additionally, we used VnMarMoT \cite{nguyenetal2017word} on the result of our word segmentation method. On the benchmark Vietnamese treebank \cite{thainp1}, we achieved result better $\text{F}_\text{1}\text{-score}$ than previous state-of-the-art result \cite{nguyenetal2017word} on Vietnamese POS Tagging task when using predicted segmentation instead of gold segmentation.
\section{Our Approach} \label{wsmethod}
In this section, we first model the word segmentation task. Next, we concentrate on the most critical part of our paper, which is the features extraction phase for the SVM classifier.

\subsection{Problem Representation}
\label{wsprorep}
In the early days of the research on Vietnamese word segmentation, the authors in \cite{diend1} considered Vietnamese word segmentation as a stochastic transduction problem. They represented the input sentence as an unweighted Finite-State Acceptor (FSA). Recently, the syllable-based and white-space-based $\text{representation}$ have been two typical ways of modeling the Vietnamese word segmentation task. The authors in \cite{nguyenetal2006} presented the syllable-based representation. In syllable-based representation, three labels B\_W, I\_W, and O\_W are used to indicate syllables that begin a word, syllables inside a word, and syllables outside a word, respectively. Syllables outside a word are punctuation marks such as full stops, commas, question marks, semicolons, and brackets. The authors  in \cite{thainp1} presented the white-space-based representation. In this representation, computers are expected to differentiate two types of white space: one appears in between two syllables of the same word, denoted by an underscore; the other separates two different words, denoted by a white space.

\begin{figure}[htb]
\centering
\includegraphics[scale=1.14]{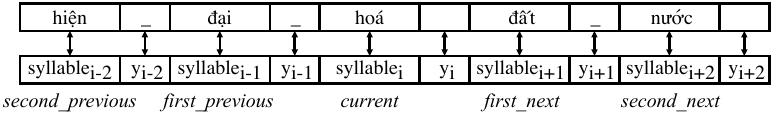}
\caption{Example of five-syllable window. In this diagram, the string ``hiện đại hóa đất nước'' ($\text{modernize}_\text{hiện\_đại\_hoá}$ $\text{country}_\text{đất\_nước}$) consisting of five syllables.}
\label{fig:1}
\end{figure}

We decided to use white-space-based representation for our Vietnamese word segmentation method because of its clarity. In our approach, we assign underscore or white space labels for each syllable from left to right of the input sentence by utilizing features in the window of five syllables from the current syllable. An example is given in Fig.~\ref{fig:1}, in which the current syllable is $\text{syllable}_\text{i}$ (``hoá''), and it needs to be classified. The gold label of $\text{syllable}_\text{i}$ is $\text{y}_\text{i}$ (white space). The five-syllable window of the current syllable contains $\text{syllable}_\text{i-2}$ (``hiện''), $\text{syllable}_\text{i-1}$ (``đại''), $\text{syllable}_\text{i}$ (``hoá''),     $\text{syllable}_\text{i+1}$ (``đất''), and $\text{syllable}_\text{i+2}$ (``nước''). Additionally, we can utilize previous labels $\text{y}_\text{i-1}$, $\text{y}_\text{i-2}$ and so on, for feature extraction of the current syllable.

\subsection{Feature Extraction}
\label{feaext}
To represent information of each syllable of the input sentence, we use the count vectorization technique. We divide the extracted features into four groups (four-vectors), which are baseline, more-than-four-syllable word, ambiguity reduction, and suffix feature. To obtain only one vector for the current syllable, we concatenated these four vectors.

We would like to introduce some utility operators and functions that we use to present feature templates for Vietnamese word segmentation. Firstly, the $\text{f}_\text{i}$ symbol represents a function that returns the lowercase-simplified form of $\text{syllable}_\text{i}$. Secondly, $\text{f}_\text{i:i+k+1}$ returns the concatenation of lowercase-simplified forms of adjacent syllables from $\text{syllable}_\text{i}$ to $\text{syllable}_\text{i+k}$ with white-space characters between them. For example given five-syllable window in Fig.~\ref{fig:1}, the value of $\text{f}_\text{i}$ symbol is ``hoá'' and value of $\text{f}_\text{i-1:i+2}$ symbol is ``đại hoá đất''. Besides, we should take syllable types into account for feature extraction. In our research, we inherit from \cite{phongnt1} four syllable types: ``lower'', ``upper'', ``all upper'', and ``other'', which correspond to the following cases: the syllable has all lowercase letters; the syllable has an upper-case initial letter; the syllable has all upper-case letters; and the syllable is a number or other things. In a similar manner as $\text{f}_\text{i}$ and $\text{f}_\text{i:i+k+1}$, we use $\text{t}_\text{i}$ and $\text{t}_\text{i:i+k+1}$ symbols for types of syllables. Lastly, range(i, i+k+1) returns the list of integers ranging $\text{from i to i+k}$ : ($\text{i, i+1, ..., i+k}$).

\subsubsection{Baseline Features}
\begin{table}[htb]
\centering
\caption{Baseline feature templates for word segmentation.}
\label{tab:1}
\begin{tabular}{|@{\hspace{0.2em}}c@{\hspace{0.2em}}|@{\hspace{0.2em}}l@{\hspace{0.2em}}|}
\hline
\textbf{No.} & \multicolumn{1}{c|}{\textbf{Templates}} \\ \hline\hline
1 & \{$\text{f}_\text{j}$ \textbf{for} j \textbf{in} \textbf{range}(i-2, i+3)\} \\ \hline
2 & \{$\text{f}_\text{j:j+2}$ \textbf{for} j \textbf{in} \textbf{range}(i-2, i+2)\} \\ \hline\hline

3 & \{(i-j) \textbf{for} j \textbf{in} \textbf{range}(i-2, i+2) \textbf{if} \textbf{inVNDict}($\text{f}_\text{j:j+2}$)\} \\ \hline
4 & \{(i-j) \textbf{for} j \textbf{in} \textbf{range}(i-2, i+1) \textbf{if} \textbf{inVNDict}($\text{f}_\text{j:j+3}$)\} \\ \hline
5 & \{(i-j) \textbf{for} j \textbf{in} \textbf{range}(i-3, i+1) \textbf{if} \textbf{inVNDict}($\text{f}_\text{j:j+4}$)\} \\ \hline\hline

6 & \{$\text{t}_\text{j:j+2}$ \textbf{for} j \textbf{in} \textbf{range}(i-2, i+2) \textbf{if} ($\text{t}_\text{j}$ \textbf{$\neq$} `LOWER' \textbf{and} $\neg$\textbf{inVNDict}($\text{f}_\text{j:j+2}$))\} \\ \hline
7 & \{$\text{t}_\text{j:j+3}$ \textbf{for} j \textbf{in} \textbf{range}(i-2, i+1) \textbf{if}  ($\text{t}_\text{j}$ \textbf{$\neq$} `LOWER' \textbf{and} $\neg$\textbf{inVNDict}($\text{f}_\text{j:j+3}$))\} \\ \hline\hline
8 & ($\text{t}_\text{i}$ \textbf{=} $\text{t}_\text{i+1}$ \textbf{=} `LOWER' \textbf{and} $\text{f}_\text{i}$ \textbf{=} $\text{f}_\text{i+1}$)? \\ \hline
9 & ($\text{t}_\text{i}$ \textbf{=} $\text{t}_\text{i+1}$ \textbf{=} `UPPER' \textbf{and} \textbf{isVNFamilyName}($\text{f}_\text{i}$))? \\ \hline
10 & ($\text{t}_\text{i}$ \textbf{=} $\text{t}_\text{i+1}$ \textbf{=} `UPPER' \textbf{and} \textbf{isVNMiddleName}($\text{f}_\text{i}$))? \\ \hline
\end{tabular}
\end{table}
\FloatBarrier

\noindent Table~\ref{tab:1} shows all feature templates of the baseline feature group. We have introduced $\text{f}_\text{i}$, $\text{f}_\text{i:i+k+1}$, $\text{t}_\text{i}$, $\text{t}_\text{i:i+k+1}$ symbols, and range(i, i+k+1) function in the last paragraph of subsection~\ref{feaext}, for convenience. \sloppy
In Table~\ref{tab:1}, inVNDict($\text{f}_\text{i:i+k+1}$) returns true if and only if $\text{f}_\text{i:i+k+1}$ is in Vietnamese word dictionary; isVNFamilyName($\text{f}_\text{i}$) returns true if and only if $\text{f}_\text{i}$ is a Vietnamese family name; isVNMiddleName($\text{f}_\text{i}$) returns true if and only if $\text{f}_\text{i}$ is a Vietnamese middle name. Notably, we used the Vietnamese words dictionary\footnote{https://github.com/datquocnguyen/RDRsegmenter/blob/master/VnVocab}, list of Vietnamese family and middle names from research of the authors in \cite{datnq1}.

In this baseline feature group, we inherit two ways of extracting feature with five-syllable window for current syllable from \cite{phongnt1}, which are the lowercase form of syllables (the first and second templates in Table~\ref{tab:1}) and syllable types (the sixth and seventh templates in Table~\ref{tab:1}). We also inherit from \cite{phongnt1} the following features: full-reduplicative word (the eighth template), Vietnamese family name (the ninth template), Vietnamese middle name (the tenth template). Additionally, we check if a conjunction of two up to four adjacent syllables in a window of seven syllables exists in the dictionary (the third, fourth, and fifth templates). These feature templates are inherited from the research of the authors in \cite{nguyenetal2006} except the fifth template.

\subsubsection{More-than-four-syllable Word Features}
\noindent We proposed this feature template based on the research of the authors in \cite{nguyenetal2006} to capture the signal of whether the center syllable is a unit of a more-than-four-syllable word. We expect the classifier can predict more-than-four-syllable words although they are rare in Vietnamese.

\begin{table}[htb]
\centering
\caption{Feature templates for capturing five up to nine syllables words.}
\label{tab:2}
\begin{tabular}{|@{\hspace{0.3em}}c@{\hspace{0.3em}}|@{\hspace{0.3em}}l@{\hspace{0.3em}}|}
\hline
\textbf{No.} & \multicolumn{1}{c|}{\textbf{Templates}} \\ \hline\hline
1 & \{(i-j) \textbf{for} j \textbf{in} \textbf{range}(i-4, i+1) \textbf{if} \textbf{inVNDict}($\text{f}_\text{j:j+5}$)\} \\ \hline
2 & \{(i-j) \textbf{for} j \textbf{in} \textbf{range}(i-5, i+1) \textbf{if} \textbf{inVNDict}($\text{f}_\text{j:j+6}$)\} \\ \hline
3 & \{(i-j) \textbf{for} j \textbf{in} \textbf{range}(i-6, i+1) \textbf{if} \textbf{inVNDict}($\text{f}_\text{j:j+7}$)\} \\ \hline
4 & \{(i-j) \textbf{for} j \textbf{in} \textbf{range}(i-7, i+1) \textbf{if} \textbf{inVNDict}($\text{f}_\text{j:j+8}$)\} \\ \hline
5 & \{(i-j) \textbf{for} j \textbf{in} \textbf{range}(i-8, i+1) \textbf{if} \textbf{inVNDict}($\text{f}_\text{j:j+9}$)\} \\ \hline
\end{tabular}
\end{table}

We recognize that words are containing up to five to nine syllables (we have shown the distribution of unique words according to lengths in Table~\ref{tab:4} of subsection~\ref{corpora}). Thus, we only take into account the concatenation of adjacent syllables with length ranging from five to nine. Lastly, we check all concatenations in the dictionary (the first, second, third, fourth, and fifth templates in Table~\ref{tab:2}).

\subsubsection{Ambiguity Reduction Features}
\label{sepsyl}

\noindent We assume that some syllables tend not to combine with other syllables in constituting a two-syllable word. For the convenience of presentation, we call the syllable with such a tendency ``a separable syllable''. We define a separable syllable as a syllable where the number of occurrences $a_i$ of one-syllable words constituted by that syllable is higher than the number of occurrences $b_i$ of more-than-one-syllable words beginning with that syllable.

\begin{figure}[htb]
\centering
\includegraphics[scale=1.16]{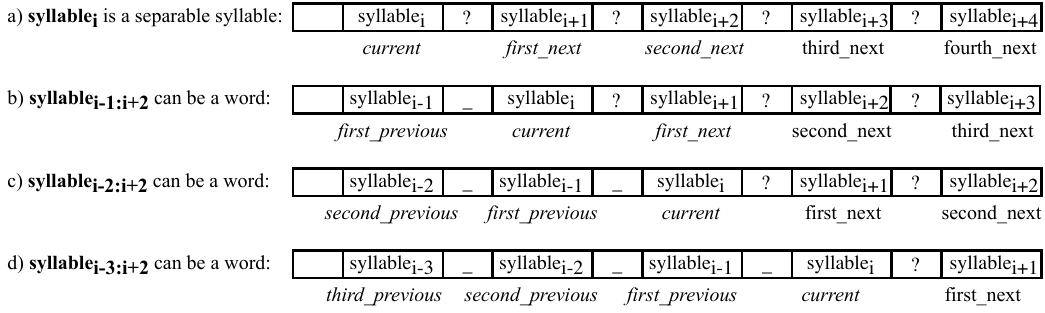}
\caption{Four situations were used in designing ambiguity reduction feature templates.}
\label{fig:2}
\end{figure}

However, we do not consider a syllable as a separable syllable if $a_i + b_i$ is not higher than the average of $a_j + b_j$ of all possible separable syllables because of we want to get rid of an uncertain separable syllable. In Vietnamese, there are some conspicuous separable syllables such as ``những'' (these), ``nhưng'' (but), ``cũng'' (also), ``đây'' (here), and ``với'' (with). The syllable ``văn'' (literature) is a non-separable syllable. For example, syllable ``văn'' usually is the first syllable of many two-syllable words such as ``văn\_bản'' (document), ``văn\_hoá'' (culture), ``văn\_sĩ'' (writer), and ``văn\_kiện'' (documentation).

\begin{table}[htb]
\centering
\caption{Feature templates in case of a current syllable is a separable syllable, and the first previous label is SPACE.}
\label{tab:3}
\begin{tabular}{|@{\hspace{0.3em}}c@{\hspace{0.3em}}|@{\hspace{0.3em}}l@{\hspace{0.3em}}|}
\hline
\textbf{No.} & \multicolumn{1}{c|}{\textbf{Templates}} \\ \hline\hline
1 & \{\textbf{inVNDict}($\text{f}_\text{j:j+2}$) \textbf{for} j \textbf{in} \textbf{range}(i, i+4)\} \\ \hline
2 & \{\textbf{inVNDict}($\text{f}_\text{j:j+3}$) \textbf{for} j \textbf{in} \textbf{range}(i, i+3)\} \\ \hline
3 & \{\textbf{inVNDict}($\text{f}_\text{j:j+4}$) \textbf{for} j \textbf{in} \textbf{range}(i, i+2)\} \\ \hline
4 & \{\textbf{inVNDict}($\text{f}_\text{j:j+5}$) \textbf{for} j \textbf{in} \textbf{range}(i, i+1)\} \\ \hline
\end{tabular}
\end{table}

The noticeable difference between our method from research of \cite{phongnt1} is that we do not use post-processing for dealing with overlap ambiguities.
We proposed a novel way of feature extraction, in which we used boolean variables to record signals of overlap ambiguity cases. In case of the current syllable is a separable syllable and the first-previous label is SPACE (as we can see in Fig.~\ref{fig:2}), we check the concatenations of lowercase-simplified forms of adjacent syllables in Vietnamese dictionary: \{$\text{f}_\text{i:i+2}$, $\text{f}_\text{i+1:i+3}$, $\text{f}_\text{i+2:i+4}$, $\text{f}_\text{i+3:i+5}$\} (the first template in Table~\ref{tab:3}); \{$\text{f}_\text{i:i+3}$, $\text{f}_\text{i+1:i+4}$, $\text{f}_\text{i+2:i+5}$\} (the second template in Table~\ref{tab:3}), \{$\text{f}_\text{i:i+4}$, $\text{f}_\text{i+1:i+5}$\} (the third template in Table~\ref{tab:3}); \{$\text{f}_\text{i:i+5}$\} (the fourth template in Table~\ref{tab:3}. In other words, we check all combinations of every two, three, four, and five adjacent syllables in a five-syllable window (as we can see in Fig.~\ref{fig:2}) in Vietnamese dictionary. This manipulation records all signals of overlap ambiguity cases, which are considered as features. We perform the same manipulation in case of $\text{syllable}_\text{i-1:i+2}$, $\text{syllable}_\text{i-2:i+2}$, and $\text{syllable}_\text{i-3:i+2}$ can be a word (described in Fig.~\ref{fig:2}).

\subsubsection{Suffix Features}

\noindent In Vietnamese, suffixes are tail-affixes (syllables or one-syllable words) that are placed after a word to create larger words \cite{nguyen1997vietnamese}. In our research, we obtain potential suffixes by statistics instead of linguistic knowledge. To obtain potential suffixes, we counted the number of occurrences of the last lower syllables in an out-of-vocabulary three-syllable or four-syllable words. However, we do not consider a syllable as a suffix if its number of occurrences is not higher than the average number of occurrences of all possible suffixes because we want to get rid of uncertain suffixes.
 
\begin{figure}[htb]
\centering
\includegraphics[scale=1.17]{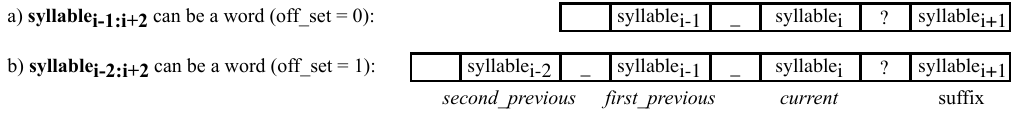}
\caption{Diagram of suffix case describes whether we choose ``underscore'' or ``space'' for current syllable.}
\label{fig:3}
\end{figure}

We design suffix features with the expectation that the classifier can predict three-syllable or four-syllable words more accurate in case of next syllable is a suffix (as we can see in Fig.~\ref{fig:3}). In other words, we want the classifier to pay special attention to the case where the next syllable is a suffix. In case the next syllable is a suffix, we derive current lowercase-simplified forms of conjunction of adjacent syllables $\text{f}_\text{i-1-off\_set:i+1}$ (the value of ``off\_set'' follows Fig.~\ref{fig:3}) as a feature for classifier. The next syllable $\text{f}_\text{i+1}$ is also treated as a feature. Finally, we derive left and right contexts of the current suffix which are $\text{f}_\text{i-2-off\_set}$, $\text{f}_\text{i-3-off\_set}$, $\text{f}_\text{i+2}$, and $\text{f}_\text{i+3}$ as features. For example, we assume that in the training set we have the string ``xây\_dựng cơ\_sở vật\_chất theo hướng \textbf{hiện\_đại\_hoá}, hoàn\_thành việc xoá lớp\_học tạm\_bợ'' (build facilities towards modernization, finish eradicating unsettled classrooms) and in the test set there is the string ``xây\_dựng nhà dân theo hướng \textbf{kiên\_cố\_hoá} để phòng\_chống lụt\_bão'' (build residential houses following solidified methods to protect against storms and floods). We also assume that in this example  ``\textbf{kiên\_cố\_hoá}'' (solidified) is out-of-vocabulary. The syllable ``\textbf{hoá}'' is a suffix in Vietnamese. In this case, the classifier can not predict the word ``\textbf{kiên\_cố\_hoá}'' in the test set because it is out-of-vocabulary. However, if we leverage the context of this suffix when training, we may predict the word ``\textbf{kiên\_cố\_hoá}''. Because it has the same the left context, uni-gram ``hướng'' and bi-gram ``theo\_hướng'', as the word ``\textbf{hiện\_đại\_hoá}'' (modernized).
\section{Experiment And Result} \label{experiment}
\subsection{Corpora}
\label{corpora}

\begin{table}[htb]
\centering
\caption{Distribution of unique words according to number of syllables in a word (\%).}
\label{tab:4}
\setlength\tabcolsep{2.7pt}
\begin{tabular}{|c|c|c|c|c|c|c|}
\hline
\multirow{2}{*}{\textbf{Corpus}} & \multicolumn{6}{c|}{\textbf{Number of syllables in a word}} \\ \cline{2-7}
 & \textbf{1} & \textbf{2} & \textbf{3} & \textbf{4} & \textbf{5-9} & \textbf{$>$9} \\ \hline\hline
VNWordSeg & 38.21 & 53.59 & 07.57 & 00.52 & 00.11 & 00.00 \\ \hline
\begin{tabular}[c]{@{}c@{}}Training dataset\\ of VLSP 2013\\ POSTag\end{tabular} & 31.66 & 58.51 & 07.33 & 02.03 & 00.45 & 00.02 \\ \hline
\begin{tabular}[c]{@{}c@{}}Training dataset\\ of VLSP 2013\\ WordSeg\end{tabular} & 36.49 & 48.92 & 11.54 & 02.63 & 00.41 & 00.01 \\ \hline
\end{tabular}
\end{table}

In our research, we compared the performance of our Vietnamese word segmentation method with published results of other well-known state-of-the-art approaches. Additionally, we studied the impact of our word segmentation method on the performance of the POS tagging task. For these purposes, we evaluated our methods on the VLSP 2013 WordSeg and VLSP 2013 POSTag corpus\footnote{http://vlsp.org.vn/vlsp2013/eval/ws-pos}, which was released for competition. Both of the two corpora are provided for research or educational purpose by the national project on Vietnamese language and speech processing VLSP\footnote{http://vlsp.org.vn}. The training dataset of VLSP 2013 WordSeg consists of 75,389 manually word-segmented sentences (approximately 23 words per sentence on average), which is part of Vietnamese treebank corpora \cite{thainp1}. The test dataset of VLSP 2013 WordSeg consists of 2,120 sentences (approximately 31 words per sentence). The training dataset of VLSP 2013 POSTag consists of 26,999 manually word-segmented sentences (about 22.5 words per sentence on average), which was collected from two sources of the national VLSP project \cite{thainp1} and the Vietnam Lexicography Center\footnote{https://www.vietlex.com}. The test dataset of VLSP 2013 POSTag consists of 2,120 sentences. Specially, we also experimented with the Vietnamese word segmentation corpus, which was provided by the authors in \cite{nguyenetal2006}. In this paper, we temporarily call this corpus ``VNWordSeg''\footnote{{https://www.jaist.ac.jp/\char`\~hieuxuan/vnwordseg/data}}. VNWordSeg consists of 7,807 manually word-segmented sentences (about 19 words per sentence on average), which was divided into 5 folds for later research \cite{nguyenetal2006}.

Table~\ref{tab:4} shows the distribution of unique words according to the number of syllables in a word in VNWordSeg, Training dataset of VLSP 2013 POSTag, and Training dataset of VLSP 2013 WordSeg. The majority of the three datasets are one- and two- syllables words. More-than-four-syllable words are rare in the three datasets. However, words containing from five to nine syllables account for the notable small ratios (0.11\%, 0.45\%, and 0.41\% in VNWordSeg, Training dataset of VLSP 2013 POSTag, and Training dataset of VLSP 2013 WordSeg, respectively). For more detail, there are 136, 305, and 321 separable syllables (described in subsection~\ref{sepsyl}) in VNWordSeg, Training dataset of VLSP 2013 POSTag, and Training dataset of VLSP 2013 WordSeg, respectively.

\subsection{Experimental Setup}
Vietnamese word segmentation has to solve the large-scale classification problem \cite{nguyenetal2006}. Therefore, we decided to use the Linear Support Vector Classification (LinearSVC) \cite{scikit11} as a tool for SVM classifier implementation. The LinearSVC on Python 3 programming language was based on LIBLINEAR  written on C programming language \cite{linearsvc}. By using LinearSVC, we tuned only one parameter, which is the penalty parameter $C$ of the error term in the SVM classifier. We chose the best value of $C$ based on the main evaluation metric $\text{F}_\text{1}\text{}$ score by using gird search experiments, in which value of $C$ can be $0.001$, $0.01$, $0.1$, $1$, $10$, or $100$.
\subsection{Feature Selection Results}

\begin{table}[H]
\centering
\caption{Our word segmentation results using 5-fold cross-validation with all combinations of features (\%). We also re-trained UETsegmenter \cite{phongnt1} and RDRsegmenter \cite{datnq1} methods with the same training datasets and testing datasets with the aim of reference.}
\label{tab:5}
\begin{tabular}{|@{\hspace{0.5em}}l@{\hspace{0.5em}}|c|c|c|c|c|c|}

\hline
\multicolumn{1}{|c|}{\multirow{5}{*}{\textbf{Prior Methods/Features}}} & \multicolumn{6}{c|}{\textbf{Corpus}} \\ \cline{2-7} 
\multicolumn{1}{|c|}{} & \multicolumn{2}{c|}{\textbf{VNWordSeg}} & \multicolumn{2}{c|}{\textbf{\begin{tabular}[c]{@{}c@{}}Training Set\\of VLSP 2013\\ POSTag\end{tabular}}} & \multicolumn{2}{c|}{\textbf{\begin{tabular}[c]{@{}c@{}}Training Set\\of VLSP 2013\\ WordSeg\end{tabular}}} \\ \cline{2-7} 
\multicolumn{1}{|c|}{}                   & \multicolumn{1}{c|}{\textbf{$C$}} & \multicolumn{1}{c|}{\textbf{$\text{F}_\text{1}\text{-score}$}} & \multicolumn{1}{c|}{\textbf{$C$}} & \multicolumn{1}{c|}{\textbf{$\text{F}_\text{1}\text{-score}$}} & \multicolumn{1}{c|}{\textbf{$C$}} & \multicolumn{1}{c|}{\textbf{$\text{F}_\text{1}\text{-score}$}} \\ \hline\hline
UETsegmenter \cite{phongnt1} &        - &    92.0986 &        - &    97.9820 &        - &    98.7954\\\hline
RDRsegmenter \cite{datnq1} &        - &  93.7811   &        - &   98.3069  &        - &  99.0726  \\\hline\hline
base &        1.0 &    94.4866 &        0.1 &    98.5080 &        0.1 &    99.2630\\ \hline\hline
base + long &        1.0 &    94.4858 &        0.1 &    98.5371 &        0.1 &    99.2762\\ \hline
base + sep &        1.0 &    94.5686 &        0.1 &    98.5647 &        0.1 &    99.2963\\ \hline
base + sfx &        1.0 &    94.4881 &        0.1 &    98.5104 &        0.1 &    99.2669\\ \hline\hline
base + long + sep &        1.0 &    94.5686 &        0.1 &    98.5848 &        0.1 &    99.3024\\ \hline
base + long + sfx &        1.0 &    94.4910 &        0.1 &    98.5434 &        0.1 &    99.2811\\ \hline
base + sep + sfx &        1.0 &    \textbf{94.5752} &        0.1 &    98.5666 &        0.1 &    99.2979\\ \hline\hline
base + long + sep + sfx &        1.0 &    94.5743 &        0.1 &    \textbf{98.5870} &        0.1 &    \textbf{99.3032}\\ \hline
\end{tabular}
\end{table}

To explore the impacts of feature groups on the performance, we conducted feature selection experiments with all combinations of features on three datasets VNWordSeg, Training dataset of VLSP 2013 POSTag, and Training dataset of VLSP 2013 WordSeg. We denoted ``base'', ``long'', ``sep'', and ``sfx'' for baseline, more-than-four-syllable word, ambiguity reduction, and suffixes feature groups, respectively.

Table~\ref{tab:5} presents feature selection results with all combinations of feature groups. More-than-four-syllable word features have impacts on the Training dataset of VLSP 2013 POSTag (0.02+\%) slightly, and Training dataset of VLSP 2013 WordSeg (0.03+\%) in comparison with the baseline groups. The ambiguity reduction features have the most substantial impacts on VNWordSeg (0.08+\%). We can also observe that the suffixes features, which have minimal impacts on three corpora (according to our experiments, there are 2, 4, and 3 suffixes on VNWordSeg, Training dataset of VLSP 2013 POSTag, and Training dataset of VLSP 2013 WordSeg, respectively).

\subsection{Main Results}

Table~\ref{tab:6} compares the Vietnamese word segmentation results of our method with results published in previous research works, using the same training and test datasets. Table~\ref{tab:6} shows that our method achieved the highest precision, recall, and $\text{F}_\text{1}\text{-score}$. Our method obtains 0.29+\% higher $\text{F}_\text{1}\text{-score}$ than RDRsegmenter \cite{datnq1}, which is the recent state-of-the-art approach. It should be noted that the results of vnTokenizer \cite{hongphuong08}, JVnSegmenter \cite{nguyenetal2006} and DongDu \cite{luu2012} were reported by the authors in \cite{phongnt1}.

\begin{table}[htb]
\centering
\caption{Word segmentation results on test dataset of VLSP 2013 WordSeg (\%).}
\label{tab:6}
\begin{tabular}{|l|c|c|c|}
\hline
\multicolumn{1}{|c|}{\textbf{Method}} & \textbf{Precision} & \textbf{Recall} & \textbf{$\text{F}_\text{1}\text{-score}$} \\ \hline\hline
vnTokenizer \cite{hongphuong08} & 96.98 &  97.69 &  97.33 \\ \hline
JVnSegmenter-Maxent \cite{nguyenetal2006} & 96.60 &  97.40 &  97.00 \\ \hline
JVnSegmenter-CRFs \cite{nguyenetal2006} &  96.63 &  97.49 & 97.06 \\ \hline
DongDu \cite{luu2012} & 96.35 & 97.46 & 96.90 \\ \hline
UETsegmenter \cite{phongnt1}  & 97.51 & 98.23 & 97.87 \\ \hline
RDRsegmenter \cite{datnq1} & 97.46 & 98.35 & 97.90 \\ \hline
Our WordSeg \{all features\} & \textbf{97.81} & \textbf{98.57} & \textbf{98.19} \\ \hline
\end{tabular}
\end{table}

Table~\ref{tab:7} shows the Vietnamese word segmentation 5-fold cross-validation results of our method with results published in previous research on the VNWordSeg corpus. Method of the authors in \cite{oanhtran10ws} had been holding the highest $\text{F}_\text{1}\text{-score}$ on VNWordSeg. However, our method obtains the highest recall score on the VNWordSeg corpus.

\begin{table}[H]
\centering
\caption{Word segmentation results using 5-fold cross-validation on VNWordSeg corpus (\%).}
\label{tab:7}
\begin{tabular}{|l|c|c|c|}
\hline
\multicolumn{1}{|c|}{\textbf{Method}} & \textbf{Precision} & \textbf{Recall} & \textbf{$\text{F}_\text{1}\text{-score}$} \\ \hline\hline
Method of the authors in \cite{nguyenetal2006} & 94.00 &  94.45 &  94.23 \\ \hline
Method of the authors in \cite{oanhtran10ws} & \textbf{96.71} &  93.89 &  \textbf{95.30} \\ \hline
Our WordSeg \{base + sep + sfx\} & 94.24 & \textbf{94.92} & 94.58 \\ \hline
\end{tabular}
\end{table}

\subsection{Analyses}

In order to analyze the word segmentation results in more detail, we computed $\text{F}_\text{1}$ score according to number of syllables in a word and three and four syllables words containing suffixes. Additionally, we also re-trained UETsegmenter \cite{phongnt1} with the Vietnamese words dictionary of RDRsegmenter \cite{datnq1} and vice versa. As we can see in Table~\ref{tab:8}, our method obtains higher $\text{F}_\text{1}$ scores than UETSegmener \cite{phongnt1}, and RDRsegmenter \cite{datnq1} on one and two syllables words (1 \& 2). On three-syllable words ($\text{3}^\text{a}$), RDRsegmenter \cite{datnq1} achieves the highest $\text{F}_\text{1}$ score. On four-syllable words ($\text{4}^\text{a}$), UETsegmenter \cite{phongnt1} achieves the highest $\text{F}_\text{1}$ score. Notably, UETsegmenter \cite{phongnt1} used another Vietnamese words dictionary\footnote{https://github.com/phongnt570/UETsegmenter/blob/master/dictionary} which contains all 7 three-and-four-syllable unknown words that they predict correctly. Besides, UETSegmener \cite{phongnt1} can not predict three syllables words containing suffixes ($\text{3}^\text{b}$) when training with the Vietnamese words dictionary of RDRsegmenter \cite{datnq1}. Therefore, we can conclude that RDRsegmenter \cite{datnq1} and our word segmentation method have not solved unknown words containing suffixes badly ($\text{3}^\text{b}$). Lastly, different from the result of UETsegmenter \cite{phongnt1} on three-syllable words ($\text{3}^\text{a}$) and RDRsegmenter \cite{datnq1} on four-syllable words ($\text{4}^\text{a}$), our result on three-syllable and words four-syllable words are not left far away by the highest result.

\begin{table}[H]
\centering
\caption{Word segmentation results ($\text{F}_\text{1}$ score) on \textbf{test dataset of VLSP 2013 WordSeg} according to number of syllables in a word (\%). For convenience, we denote three and four syllables unknown words containing suffixes by \textbf{$\text{3}^\text{b}$} and \textbf{$\text{4}^\text{b}$} (unknown words are detected by checking in the Vietnamese words dictionary of RDRsegmenter \cite{datnq1}). And conversely, we use \textbf{$\text{3}^\text{a}$} and \textbf{$\text{4}^\text{a}$}, indicating three and four syllables words which are not \textbf{$\text{3}^\text{b}$} or \textbf{$\text{4}^\text{b}$}. Notably, we temporarily use \textbf{UETws}, \textbf{RDRws}, and \textbf{UITws} as abbreviations for \textbf{UETsegmenter} \cite{phongnt1}, \textbf{RDRsegmenter} \cite{datnq1}, and \textbf{our word segmentation method using all features}. We also provide proportions of words (\%) in parentheses.}
\label{tab:8}
\resizebox{\textwidth}{!}{%
\begin{tabular}{|c|l|c|c|c|c|c|c|c|c|}
\hline
\multirow{3}{*}{\textbf{\begin{tabular}[c]{@{}c@{}}Vietnamese\\ Dictionary\\ Resource\end{tabular}}} & \multicolumn{1}{c|}{\multirow{3}{*}{\textbf{Method}}} & \multicolumn{7}{c|}{\textbf{Number of syllables in a word}} & \multirow{3}{*}{\textbf{Total}} \\ \cline{3-9}
 & \multicolumn{1}{c|}{} & \textbf{\begin{tabular}[c]{@{}c@{}}1\\ (57.75)\end{tabular}} & \textbf{\begin{tabular}[c]{@{}c@{}}2\\ (40.42)\end{tabular}} & \textbf{\begin{tabular}[c]{@{}c@{}}$\text{3}^\text{a}$\\ (00.74)\end{tabular}} & \textbf{\begin{tabular}[c]{@{}c@{}}$\text{3}^\text{b}$\\ (00.13)\end{tabular}} & \textbf{\begin{tabular}[c]{@{}c@{}}$\text{4}^\text{a}$\\ (00.68)\end{tabular}} & \textbf{\begin{tabular}[c]{@{}c@{}}$\text{4}^\text{b}$\\ (00.05)\end{tabular}} & \textbf{\begin{tabular}[c]{@{}c@{}}5-9\\ (00.22)\end{tabular}} & \\ \hline\hline
\multirow{3}{*}{\begin{tabular}[c]{@{}c@{}}UETws\\ \cite{phongnt1}\end{tabular}} & UETws \cite{phongnt1} & 98.46 & \textbf{97.97} & 79.96 & \textbf{89.74} & \textbf{78.62} & \textbf{100.00} & 21.30 & 97.87 \\ \cline{2-10} 
 & RDRws \cite{datnq1} & 98.37 & 97.68 & 85.41 & 89.03 & 74.23 & \textbf{100.00} & 23.60 & 97.74 \\ \cline{2-10} 
 & UITws & \textbf{98.59} & 97.96 & \textbf{85.77} & \textbf{89.74} & 77.26 & \textbf{100.00} & \textbf{34.02} & \textbf{98.01} \\ \hline\hline
\multirow{3}{*}{\begin{tabular}[c]{@{}c@{}}RDRws\\ \cite{datnq1}\end{tabular}} & UETws \cite{phongnt1} & 98.47 & 97.90 & 80.40 & 0.00 & \textbf{79.51} & \textbf{26.32} & 34.97 & 97.79 \\ \cline{2-10} 
 & RDRws \cite{datnq1} & 98.57 & 97.85 & \textbf{86.30} & 79.19 & 75.74 & 0.00 & 23.60 & 97.90 \\ \cline{2-10} 
 & UITws & \textbf{98.82} & \textbf{98.14} & 85.23 & \textbf{80.20} & 78.60 & 0.00 & \textbf{46.83} & \textbf{98.19} \\ \hline
\end{tabular}%
}
\end{table}

Lastly, Table~\ref{tab:9} shows POS tagging performance on the test dataset of VLSP 2013 POSTag with the predicted word segmentation. We re-trained the UETsegmenter tool on VLSP 2013 POSTag. Our Vietnamese word segmentation method has helped VnMarMot \cite{nguyenetal2017word} of increase in performance on VLSP 2013 POSTag with 0.3+\% improvement of $\text{F}_\text{1}$ score by comparing with (VnMarMoT \cite{nguyenetal2017word} using RDRsegmenter \cite{datnq1}) approach.

\begin{table}[H]
\centering
\caption{POS Tagging performance with predicted word segmentation on test dataset of VLSP 2013 POSTag (\%).}
\label{tab:9}
\begin{tabular}{|@{\hspace{0.25em}}l@{\hspace{0.25em}}|c|c|}
\hline
\multicolumn{1}{|c|}{\multirow{2}{*}{\textbf{Method}}} & \multicolumn{2}{c|}{\textbf{$\text{F}_\text{1}\text{-score}$}} \\ \cline{2-3} 
\multicolumn{1}{|c|}{} & \textbf{WordSeg} & \multicolumn{1}{l|}{\textbf{POSTag}} \\ \hline\hline
RDRPOSTagger with RDRsegmenter \cite{nguyenetal2017word} & 97.75 & 93.39 \\ \hline
(BiLSTM-CRF + CNN-char) with RDRsegmenter \cite{nguyenetal2017word} & 97.75 & 93.55 \\ \hline
VnMarMoT with RDRsegmenter \cite{nguyenetal2017word} & 97.75 & 93.96 \\ \hline
VnMarMoT \cite{nguyenetal2017word} with Our WordSeg \{all features\} & \textbf{98.06} & \textbf{94.27} \\ \hline
\end{tabular}
\end{table}
\section{Conclusion and Future Work} \label{conclusion}
In this paper, we propose a novel feature-based method using the SVM classifier for Vietnamese word segmentation. Overlap ambiguity and unknown words containing suffixes phenomena are real challenges in Vietnamese word segmentation. We prove that our proposed features, ambiguity reduction and suffix-capturing features, help to improve the performance of word segmentation. Experiments on the benchmark Vietnamese datasets show that our method obtains a higher $\text{F}_\text{1}\text{-score}$ score than state-of-the-art approaches. Finally, according to the experimental results, our Vietnamese word segmentation method has a positive impact on Vietnamese POS tagging. However, the greatest weakness of our ambiguity reduction and suffix features is that we do not care about parts-of-speech information. Therefore, we are planning to refer to the ambiguity solving method of the authors in \cite{ducpd1} for our further research. Our code is open-source and available at {\tt https://github.com/ngannlt/UITws-v1}.

\section*{Acknowledgment}
This research is funded by University of Information Technology-Vietnam \mbox{National} University HoChiMinh City under grant number D1-2019-16.

\addcontentsline{toc}{section}{References}
\bibliographystyle{splncs04}
\bibliography{bibliography}

\end{document}